\documentclass[10pt,twocolumn,letterpaper]{article}

\usepackage[review,applications]{wacv}
\usepackage{times}
\usepackage{epsfig}
\usepackage{graphicx}
\usepackage{graphics}
\usepackage{amsmath}
\usepackage{amssymb}
\usepackage{amsfonts}
\usepackage{multirow}
\usepackage{color}
\usepackage[table]{xcolor}

\usepackage{arydshln}
\usepackage{bm}
\usepackage{bbm}
\usepackage{colortbl}
\usepackage{hyperref}
\hypersetup{%
 colorlinks=true}

\usepackage[capitalize]{cleveref}
\crefname{section}{Sec.}{Secs.}
\Crefname{section}{Section}{Sections}
\Crefname{table}{Table}{Tables}
\crefname{table}{Tab.}{Tabs.}

\def\wacvPaperID{1465} 
\def\confName{WACV}
\def\confYear{2025}

\begin{document}

\title{Ordinal Multiple-instance Learning for Ulcerative Colitis Severity Estimation\\with Selective Aggregated Transformer}

\author{Kaito Shiku$^{1}$
\and
Kazuya  Nishimura$^{2}$
\and
Daiki Suehiro$^{1}$
\and
Kiyohito Tanaka$^{3}$
\and
Ryoma Bise$^{1}$
\and
{$^{1}$ Kyushu University}
\and
{$^{2}$ National Cancer Center Japan}
\and
{$^{3}$ Kyoto Second Red Cross Hospital}
\\
{\tt\small kaito.shiku@human.ait.kyushu-u.ac.jp}
}

\maketitle


\begin{abstract}
\vspace{-3mm}
Patient-level diagnosis of severity in ulcerative colitis (UC) is common in real clinical settings, where the most severe score in a patient is recorded.
However, previous UC classification methods ({\it i.e.,} image-level estimation) mainly assumed the input was a single image. Thus, these methods can not utilize severity labels recorded in real clinical settings.
In this paper, we propose a patient-level severity estimation method by a transformer with selective aggregator tokens, where a severity label is estimated from multiple images taken from a patient, similar to a clinical setting.
Our method can effectively aggregate features of severe parts from a set of images captured in each patient, and it facilitates improving the discriminative ability between adjacent severity classes.
Experiments demonstrate the effectiveness of the proposed method on two datasets compared with the state-of-the-art MIL methods.
Moreover, we evaluated our method in real clinical settings and confirmed that our method outperformed the previous image-level methods.
The code is publicly available at \url{https://github.com/Shiku-Kaito/Ordinal-Multiple-instance-Learning-for-Ulcerative-Colitis-Severity-Estimation}.

\vspace{-4mm}

\end{abstract}

\section{Introduction}

\begin{figure}
      \centering
        \includegraphics[width=1.0\linewidth]{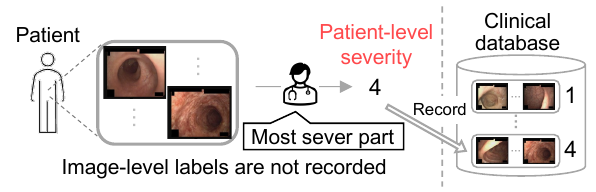}
        \vspace{-6mm}
        \caption{{\bf Patient-level diagnosis of severity in ulcerative colitis (UC).}
        In real clinical settings, approximately 20 to 40 endoscopic images are captured from each patient, and the severity score for the most severe area of the images is recorded as the patient-level diagnosis.
        In this process, the severity score for each image is not recorded.}     
        \label{fig:intro}
        \vspace{-5mm}
\end{figure}

Patient-level severity in ulcerative colitis (UC) diagnosis
has been recorded in real clinical settings, creating a valuable and extensive dataset, as shown in Figure~\ref{fig:intro}.
For example, in the UC diagnosis, approximately 20 to 40 endoscopic images are captured for each patient. Only the severity score for the most severe area of the images is recorded as the patient-level diagnosis, while detailed information for individual images ({\it i.e.,} image-level severity score) is not provided.

The goal of this paper is to develop a patient-level UC severity estimation method that utilizes accumulated patient-level diagnostic records for training, without relying on image-level labels.
The patient-level estimation can be formulated as a multiple-instance learning (MIL) problem~\cite{ilse2018attention,javed2022additive,li2021dual,lin2023interventional,schwab2022automatic,shao2021transmil,wang2018revisiting}, which estimates severity labels from a bag consisting of a set of instances ({\it i.e.,} endoscopic images). We emphasize that our method only uses existing data that can be obtained from diagnoses recorded and accumulated in real clinical settings, without requiring additional annotations ({\it e.g.,} image-level annotations).

The specific problems in the patient-level UC severity estimation are that severity scores have an ordinal relationship, and the severity for the most severe area of the images is recorded as the patient-level diagnosis (called max severity estimation in ordinal MIL).
This task makes it challenging to obtain discriminative bag-level features between neighboring classes. 
This challenge arises because MIL methods generally require aggregating instance features within a bag to obtain bag-level features, and instance features with lower severity than the patient-level severity may also be aggregated. Consequently, bag-level features with high severity are influenced by those with lower severity, reducing the distinction between classes.

\begin{figure}
      \centering
        \includegraphics[width=1.0\linewidth]{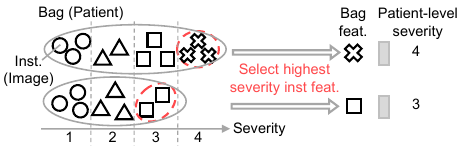}
        \vspace{-6mm}
        \caption{
        {\bf Ideal feature aggregation for max severity estimation in ordinal MIL.}
        The patient-level severity is defined based on the label of the most severe instance within the bag.
        In this setting, the ideal aggregation involves aggregating features from the instances with the highest severity label within a bag.
        }
        \vspace{-4mm}
        \label{fig:Ideal_aggregation}
\end{figure}

Ideal feature aggregation for max severity estimation in ordinal MIL involves aggregating features only from instances with the highest severity label within a bag, as shown in Figure~\ref{fig:Ideal_aggregation}.
Realizing this approach reduces the negative impact of instance features with lower severity.

While conventional attention MIL methods~\cite{ilse2018attention,javed2022additive,shao2021transmil} reduce their influence by weighting instance features using attention, 
they also have limitations in assigning high attention not only to the highest severity instances but also to others.
This is because conventional methods generate a single bag-level feature to classify all severity classes; the same attention strategy should be used for bags with different severity 1, 2, 3, and 4. The detailed analysis will be discussed in the experimental section.

\begin{figure}
      \centering
        \includegraphics[width=0.85\linewidth]{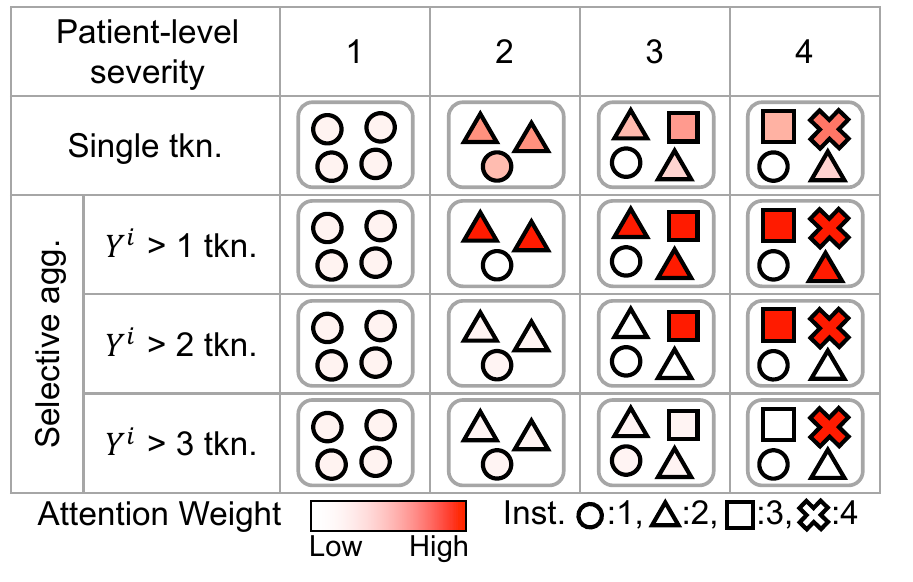}
        \vspace{-3mm}
        \caption{{\bf Illustration of the effect of our selective aggregation.}
        A previous single token may give attention for many instances, not only the severe ones, resulting in un-discriminative bag-level features.
        In our approach, each selective aggregator token corresponds to a different severity level, with the $k$-th token aggregating instance features that satisfy $\{Y^i>k\}$.
Each token can effectively aggregate the severe instance features for each severity level of bags, leading to the production of discriminative bag-level features to distinguish between adjacent severity classes.
        }
        \label{fig:concept}
        \vspace{-3mm}
\end{figure}

To address the difficulties, we propose a Selective Aggregated Transformer for Ordinary MIL (SATOMIL) that selectively aggregates the highest severity instance features from a set of patient images.
Specifically, we introduce $k-1$ selective aggregator tokens. 
Each token is responsible for aggregating instance features from classes above a certain threshold, and the $k$-th token selectively aggregates only those instance features satisfying $Y^i>k$.
In the example, as shown in Figure~\ref{fig:concept}, by estimating high attention only on instances of class 4 that satisfy $Y^i > 3$, it is possible to obtain discriminative bag-level features for adjacent classes.
By introducing the selective aggregator tokens for all $k=1,\ldots, K-1$, we can obtain the suitable bag-level features for each severity class and significantly improve the accuracy.

In the experiments using two endoscopic image datasets, SATOMIL outperformed the state-of-the-art MIL methods in patient-level severity estimation. In addition, an ablation study demonstrated that our method improved the accuracy compared to the method that naively introduced the ordinal classification algorithm into the MIL baseline to consider the ordinality of the classes.
Furthermore, our method trained with patient-level annotation achieved higher accuracy than the image-level severity estimation method trained with the supervised data for image-level severity and aggregates the estimated labels of individual images.

Our main contribution is summarized as follows:

\begin{itemize}
\item We propose a framework for max severity estimation in ordinal MIL that only uses existing data that can be obtained from diagnoses recorded in real clinical settings, without requiring additional annotation.
\item We propose a Selective Aggregated Transformer designed to address max severity estimation in ordinal MIL. This model focuses on aggregating features from 
severe 
instances. We introduce selective aggregator tokens to perform aggregation through attention mechanisms.
\item The experimental results using real clinical endoscopic data demonstrated the effectiveness of the proposed method. The proposed method outperformed the state-of-the-art MIL methods.
\end{itemize}

\section{Related work}
\noindent
{\bf Automated image-level UC severity estimation:}
Automated UC severity estimation methods for individual images have been proposed with deep neural networks trained with image-level annotations \cite{kadota2022automatic, kadota2022deep,stidham2019performance, takenaka2020development, isbi_takezaki,polat2022class}. The main paradigm of the UC severity estimation is to utilize the ordinal relationship of labels. For example, Kadota {\it et al.} \cite{kadota2022automatic} have utilized ordinality by jointly learning regression and ranking tasks. 
However, since these methods require image-level annotation, they could not benefit from the clinical database that is recorded in clinical diagnosis.
The clinical database only contains patient-level annotations; therefore, a patient-level estimation method is more label-efficient.

Note that patient-level estimation is not only applicable to UC diagnosis but also to many endoscopic diagnoses, which may not have public datasets. Therefore, it is important to develop robust patient-level estimation methods.


\noindent
{\bf Automated patient-level UC severity estimation :}
Automated patient-level UC severity estimation has not been thoroughly studied, and there are few related methods.
Schwab et al.~\cite{schwab2022automatic} simply introduced the K-rank label estimation~\cite{cao2020rank,niu2016ordinal} into MIL to account for ordinal classification. However, this method is based on output aggregation and does not aggregate instance features with the highest severity labels. For instance, in their paper, the K-rank-based method showed almost the same performance as the regression-based MIL method.
Our method introduces selective aggregator tokens to effectively select instance features with the maximum severity in a bag. This approach is significantly different from the aim of the original K-rank methods.

\noindent
{\bf Multiple-instance learning:}
Multiple-instance learning (MIL)~\cite{ilse2018attention,javed2022additive,li2021dual,lin2023interventional,schwab2022automatic,shao2021transmil,wang2018revisiting} aims to estimate bag-level labels for bags composed of multiple instances (i.e., images).
MIL methods are divided into output aggregation methods ~\cite{ramon2000multi,wang2018revisiting} and feature aggregation methods ~\cite{ilse2018attention,Ilse2020DeepMI,li2021dual,lin2023interventional,Pinheiro2015,shao2021transmil,yu2021mil,bi2021local}.
Output aggregation first estimates the class label for each instance and aggregates these outputs. Feature aggregation method aggregates instance features to obtain bag-level features.

Although one of the most straightforward approaches is Ouptput+Max~\cite{wang2018revisiting}, which aggregates the output scores by max pooling, this method does not work. 
This is because the loss is only propagated to an instance that achieves the maximum within a bag; thus, the loss disappears if one instance shares the correct label~\cite{ilse2018attention}.
In recent research, feature aggregation methods have become mainstream due to its advantage for feature representation.
However, conventional feature aggregation methods~\cite{Ilse2020DeepMI} tend to aggregate all instance features as a bag-level feature, which is unsuitable for our task.

Recently, Transformer-based MIL methods~\cite{javed2022additive,lin2023interventional,shao2021transmil,yu2021mil} have been proposed, that aggregate instance features with a single class token. Estimating proper attention for instances with a single class token becomes challenging when there is an ordinal relationship among instance classes, as shown in Figure~\ref{fig:concept} (Top); which may give attention for many instances, not only the severe ones (please refer the introduction for detailed discussions).
In previous research, no studies have focused on aggregation methods to address the specific problems of max severity estimation in ordinal MIL. 

\noindent
{\bf Ordinal classification:}
Ordinal classification methods account for ordinal relationships between classes~\cite{cao2020rank,diaz2019soft,guo2013joint,li2021learning,niu2016ordinal,Wang_Jiang_Yin_Cheng_Ge_Gu_2023}.
The basic idea of ordinal classification is to avoid misclassifying into classes that are far apart in order. It imposes a large penalty when the estimated class is far from the ground truth and a small penalty when the estimated class is close to the ground truth.
Ordinal classification methods are widely divided into regression~\cite{guo2013joint}, classification~\cite{diaz2019soft}, and K-rank algorithms~\cite{cao2020rank,niu2016ordinal}.
Some methods impose constraints on the ordinal distribution within the feature space~\cite{Wang_Jiang_Yin_Cheng_Ge_Gu_2023}.
These methods are typically designed to learn from labels assigned at the image level. Therefore, the aggregation of image features, which is an important process in ordinal MIL, is not performed.

\begin{figure*}[t]
 \begin{center}
    \includegraphics[width=0.9\textwidth]{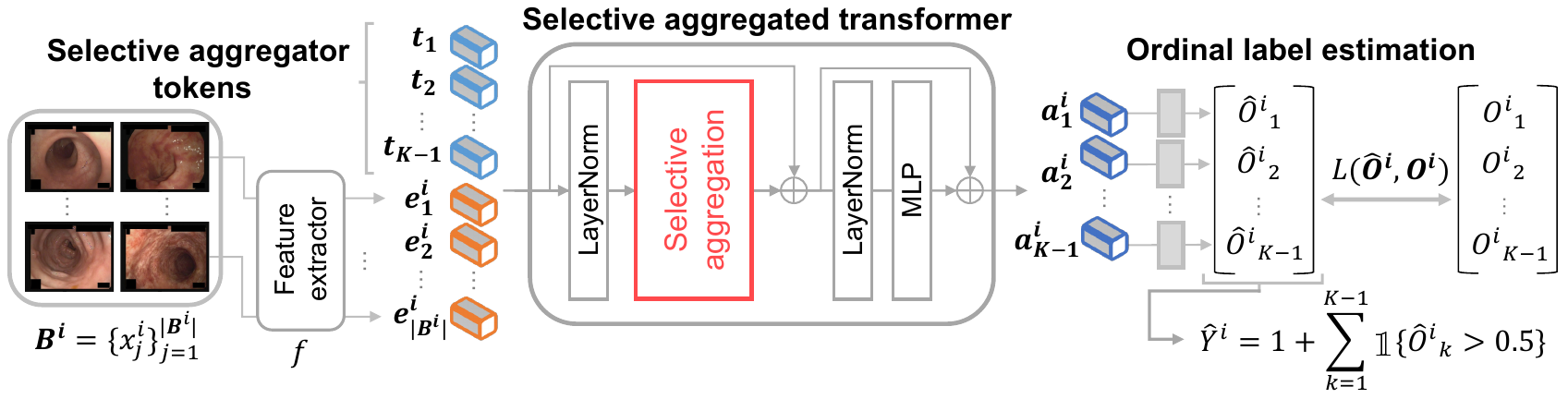}
    \vspace{-3mm}
    \caption{{\bf Overview of Selective Aggregated Transformer.} First, instance features $\bm{e}_j^i$ are extracted from each image. Then, the features are aggregated by selective aggregator tokens to obtain the bag-level features $\bm{a}_1^i,...,\bm{a}_{K-1}^i$, which focus on each class boundary. Finally, rank predictions are obtained by applying binary classifiers for each aggregated feature.}
    \vspace{-6mm}
    \label{fig:overview}
 \end{center}
\end{figure*}

\section{Patient-level max severity estimation in Ordinal Multiple-Instance Learning.}
\subsection{Problem setup}
Given a set of endoscopic images for $i$-th patient ({\it i.e.,} bag) $\mathcal{B}^i = \{\bm{x}_{j}^i \}_{j=1}^{|\mathcal{B}^i|}$, where $\bm{x}_j^i$ is a colonoscopy image ({\it i.e.,} instance), and $|\mathcal{B}^i|$ is the total number of images, our aim is to estimate severity label $Y^i$ for the patient. 
Here, the severity labels belong to $K$ classes $Y^i \in \{1, 2,\ldots,K\}$, where each class has ordinal relationships $K \succ K-1,...,\succ 1$.
Note that patient-level severity is defined as the highest severity among all the individual images of the patient; however, the severity of each individual image is unknown, and only the patient-level severity is provided in the training data because the patient-level severity has been reported only in clinical diagnosis.

The task is one of ordinal multiple-instance learning setups in which the target label has ordinal relationships. 

\subsection{Motivation of our approach}
Our approach employs a feature aggregation strategy. 
A straightforward method for estimating the highest severity in a bag is output max~\cite{wang2018revisiting}  aggregation, which selects the maximum score from individual estimations in Multiple Instance Learning (MIL).
However, it is widely recognized that output aggregation does not perform well for feature representation in MIL \cite{ilse2018attention}. 
This is because the loss is only propagated to an instance that achieves the maximum within a bag; thus, the loss disappears if one instance shares the correct label. In contrast, a feature aggregation strategy can give a loss to instances with the maximum label and other instances, and thus, feature aggregation is known to be superior to output aggregation~\cite{ilse2018attention,li2021dual,lin2023interventional,shao2021transmil}.
Therefore, we use a feature aggregation strategy.

We introduce multiple selective aggregator tokens into a transformer to address max severity estimation in ordinal MIL.
Each selective aggregator token corresponds to a different severity level, with the $k$-th token aggregating instance features that satisfy $\{Y^i>k\},(k=1,\ldots,K)$.
Each token can effectively aggregate the severe instance features for each severity level of bags.
Consequently, each token can obtain discriminative bag-level features to distinguish between adjacent severity classes.

Figure~\ref{fig:concept} shows examples of the role of each token, where there are four severity classes. In this case, the transformer has three selective aggregator tokens.
Token $\{Y^i>1\}$ assigns equal-level attention for all instances in a bag with severity 1 and higher attention for instance features whose severity is larger than 1 in other bags.
Similarly, token $\{Y^i>2\}$ assigns the same-level attention for all instances in a bag with severity 1 and 2 and higher attention to instance features whose severity is larger than 2 in bags with severity 3 and 4.

Note that the aim of introducing K-rank tokens differs significantly from the traditional K-rank algorithm~\cite{cao2020rank,niu2016ordinal}. Traditional K-rank approaches multi-label learning by estimating conditions for each K-rank and integrating these for severity-level estimation to introduce ordinal class distances into the training loss.
In contrast, our method aims to selectively aggregate instance features to obtain a bag-level feature in MIL. The details of the method are described below.

\begin{figure}
      \centering
        \includegraphics[width=1.0\linewidth]{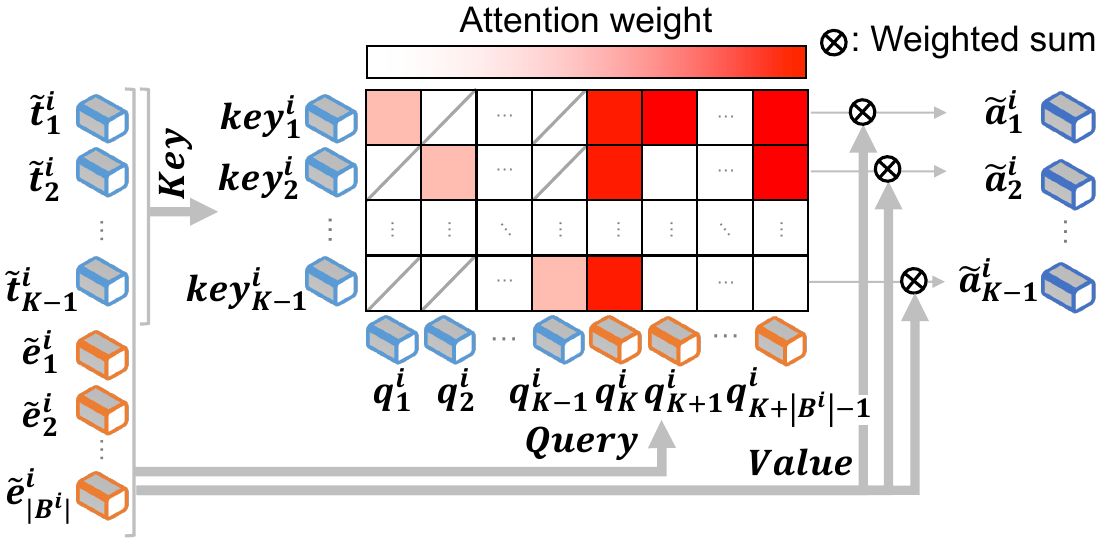}
        \vspace{-5mm}
        \caption{{\bf Selective aggregation.} The $k$-th selective aggregator token $\tilde{\bm{t}}_j^{i}$ aggregates the instance features $\tilde{\bm{e}}_j^{i}$ to extract the discriminative bag-level feature. To effectively discriminate $Y^i>{k}$, instance features above $k$ should be aggregated while ignoring instances less than or equal to $k$. Therefore, training makes attention to severe instances above $k$ higher (red), and otherwise lower (white).}
        \vspace{-4mm}
        \label{fig:selective_aggregation}
\end{figure}

\subsection{Selective aggregated transformer}
Figure~\ref{fig:overview} shows an overview of the Selective Aggregated Transformer for the Ordinary MIL (SATOMIL), which selectively aggregates instance features while focusing on severe instances.
First, given a bag, the instance feature for each image $\bm{x}_j^i$ is extracted by the feature extractor $f$, and these features are inputted into a transformer with selective aggregator tokens $\{\bm{t}_1,\ldots,\bm{t}_{K-1}\}$, which is trainable in training. 
Then, we aggregate the features $\bm{e}_j^i = f(\bm{x}_j^i)$ by selective aggregator tokens to obtain the bag-level features $\bm{a}_1^i,...,\bm{a}_{K-1}^i$ focused on each class boundary.
Finally, a patient-level severity label $\hat{Y}^i$ is estimated in a K-rank classification manner.

The severity class $Y^i$ is extended into a series of binary classification labels $[O_{1}^i, \ldots ,O_{K-1}^i ]$, similar to the K-rank algorithm~\cite{cao2020rank,niu2016ordinal}. Here, $O_{k}^i \in \{0,1\}$ indicates whether the label $Y^i$ of $i$-th bag exceeds class $k$, where $O_{k}^i = \mathbbm{1}\{Y^i>{k}\}$ and the indicator function $\mathbbm{1}\{ \cdot \}$ returns 1 if the condition is true and 0 otherwise.

These multi-labels are estimated using individual selective aggregator tokens that aggregate instance features; the label ${\hat{O}_{k}^i}$ is estimated using the $k$-th token.
The feature aggregation is performed based on self-attention in the selective aggregated transformer, as shown in Figure~\ref{fig:overview}.
The set of the extracted features for individual instances $\mathcal{T}_{p}^i = \{\bm{e}_j^i\}_{j=1}^{|\mathcal{B}^i|}$ and the set of selective aggregator tokens $\mathcal{T}_{s} = \{ \bm{t}_1, ..., \bm{t}_{K-1} \}$ are inputted into the selective aggregated transformer (SAT).
The SAT outputs the aggregated bag-level features $\{\bm{a}_{1}^i,\ldots,\bm{a}_{K-1}^i\}$ for estimating the series of binary labels $\{O_{k}^i\}$ as follows:
\begin{equation}
    \bm{a}_{1}^i,...,\bm{a}_{K-1}^i = \mathrm{SAT}(\mathcal{T}_s, \mathcal{T}_p^i).
\end{equation}

As shown in Figure~\ref{fig:selective_aggregation}, selective aggregation in SAT aggregates the instance features by using selective aggregator tokens. Here, we denote the set of instance features and selective aggregator tokens following the first LayerNorm layer in Figure~\ref{fig:overview} as $\tilde{\mathcal{T}}_{p}^{i} = \{\tilde{\bm{e}}_j^{i}\}_{j=1}^{|\mathcal{B}^i|}$ and $\tilde{\mathcal{T}}_{s}^{i} = \{ \tilde{\bm{t}}_1^{i}, ..., \tilde{\bm{t}}_{K-1}^{i} \}$, respectively.
The $k$-th selective aggregator token $\tilde{\bm{t}}_k^{i}$ aggregates the instance features in the $i$-th bag to extract the bag-level feature for estimating $O_{k}^i = \mathbbm{1}\{Y^i>{k}\}$. 
The queries, keys, and values are extracted from the selective aggregator tokens and instance features, denoting $\bm{\mathrm{key}}_1^i,...,\bm{\mathrm{key}}_{K-1}^i=Key(\tilde{\mathcal{T}}_{s}^{i})$, $\bm{q}_1^i,...,\bm{q}_{K+|\mathcal{B}^i|-1}^i=Query(\tilde{\mathcal{T}}_{s}^{i}, \tilde{\mathcal{T}}_{p}^{i})$, and $\bm{v}_1^i,...,\bm{v}_{K+|\mathcal{B}^i|-1}^i=Value(\tilde{\mathcal{T}}_{s}^{i}, \tilde{\mathcal{T}}_{p}^{i})$, respectively. 
The scale-dot attention is calculated between the key $\bm{\mathrm{key}}_k^i$ and the query 
$\bm{q}_l^i$ 
by the dot similarity $s_{kl}^i = \frac{\bm{q}_l^i \cdot \bm{\mathrm{key}}^i_k}{\sqrt{d}}$, where $d$ is dimension.
Since the goal of this self-attention is instance feature aggregation, a mask is used to ignore attention between selective aggregator tokens.
Consequently, the aggregated bag-level feature is calculated as follows:
$\tilde{\bm{a}}_k^{i} = \sum_{l=1}^{K+|\mathcal{B}^i|-1} {m}_{kl} \cdot \frac{\exp \left( s_{
kl}^i \right)}{\sum_{r=1}^{K+|\mathcal{B}^i|-1} \exp \left(s_{kr}^i \right)} \cdot \bm{v}^i_l$
, where $m_{kl}$ is an element of the mask that is 0 for other selective aggregation tokens and 1 otherwise.
After selective aggregation, $\bm{a}_{k}^i$ is obtained from $\tilde{\bm{a}}_k^{i}$ through a second LayerNorm and MLP.
To effectively discriminate $Y^i>{k}$, instance features above $k$ should be aggregated while ignoring instances less than or equal to $k$. Therefore, training makes attention to severe instances above $k$ higher, and otherwise lower.
This attention process is performed for all tokens.

The rank prediction $\hat{O}^i_{k}$ is obtained from the aggregated features $\bm{a}_{k}^i$ using the binary classifier $g_k$, $[\hat{O}_{1}^i,...,\hat{O}_{K-1}^i] = [g_{1}( \bm{a}_{1}^i),...,g_{K-1}(\bm{a}_{K-1}^i)]$.
The model is trained with the binary cross entropy loss for each prediction $\hat{O}^i_{k}$:

\vspace{-4mm}
\begin{equation}
\begin{split}
L(O^i, \hat{O}^i) = - \sum_{i=1}^n \sum_{k=1}^{K-1} &\left( O_{k}^i  \log(\hat{O}_{k}^i) \right. \\
&\left. + (1 - O_{k}^i) \log(1 - \hat{O}_{k}^i) \right).
\end{split}
\end{equation}
The loss function imposes a large penalty when the estimated class is far away from the ground truth and a small penalty when the estimated class is close to the ground truth. It reflects the ordinal relationship of the labels.




In the inference phase, the final severity prediction $\hat{Y}^i$ is obtained from the estimation of each binary classifier $\hat{O}_{k}^i$ with 0.5 thresholding:
\vspace{-1mm}
\begin{equation}
\vspace{-1mm}
\begin{split}
&\hat{Y}^i = 1 + \sum_{k=1}^{K-1} \mathbbm{1}\{\hat{O}_{k}^i  > 0.5\}.
\end{split}
\end{equation}

\begin{table*}
    \centering
        \caption{{\bf Comparison with MIL methods on LIMUC and Private dataset.} Best performances are bold.
        }
        \vspace{-2mm}
        \scalebox{0.97}{
        \begin{tabular}{c || ccc ||ccc}  \hline
         \multirow{2}{*}{Method} &  \multicolumn{3}{c||}{LIMUC} & \multicolumn{3}{c}{Private} \\
            &  Accuracy & Kappa & Macro-f1 & Accuracy & Kappa & Macro-f1\\ \hline \hline
            Output+Mean~\cite{wang2018revisiting}  & 0.465 & 0.482 & 0.432 & 0.518 & 0.502 & 0.485\\
        Output+Max~\cite{wang2018revisiting} & 0.465 & 0.574 & 0.446 & 0.464 & 0.457 & 0.338\\
            Feature+Mean~\cite{Ilse2020DeepMI}& 0.534 & 0.624 & 0.482 & 0.624 & 0.682 & 0.596\\
            Feature+Max~\cite{Ilse2020DeepMI} & 0.596 & 0.694 & 0.525 & 0.582 & 0.587 & 0.461\\
            Feature+Attention~\cite{ilse2018attention}&  0.635 & 0.751 & 0.593 & 0.634 & 0.677 & 0.589\\
            Transformer& 0.665 & 0.813 & 0.641 & 0.634 & 0.730 & 0.608\\
            DSMIL~\cite{li2021dual} & 0.603 & 0.744 & 0.578 & 0.634 & 0.681 & 0.598\\ 
            AdditiveTransMIL~\cite{javed2022additive}& 0.561 & 0.698 & 0.543 & 0.567 & 0.589 & 0.515\\ 
            IBMIL~\cite{lin2023interventional} & 0.615 & 0.731 & 0.578 & 0.629 & 0.635 & 0.541\\   
            K-rank MIL~\cite{schwab2022automatic} & 0.633 & 0.797 & 0.601 & 0.600 & 0.683 & 0.575\\
                \rowcolor{gray!15}
            Ours & \bf{0.690} & \bf{0.826} & \bf{0.674} & \bf{0.683} & \bf{0.774} & \bf{0.646}\\ \hline
    
        
        \end{tabular}
        }
        \vspace{0mm}
        \label{tab:milcomparison}
\end{table*}

\begin{table*}
    \centering
        \caption{{\bf Comparison with ordinal classification methods on LIMUC and Private dataset.}  Best performances are bold.
        We introduced five ordinal classification methods into the Transformer, which is the second-best among MIL methods.}    
        \vspace{-2mm}
        \scalebox{0.97}{
        \begin{tabular}{c || ccc ||ccc}  \hline
         \multirow{2}{*}{Method} &  \multicolumn{3}{c||}{LIMUC} & \multicolumn{3}{c}{Private} \\
            &  Accuracy & Kappa & Macro-f1 & Accuracy & Kappa & Macro-f1\\ \hline \hline
    
            Transformer regression~\cite{guo2013joint} & 0.623 & 0.792 & 0.609 & 0.650 & 0.730 & 0.585\\
        Transformer soft label~\cite{diaz2019soft} & 0.651 & 0.808 & 0.634 & 0.665 & 0.740 & 0.640\\
            Transformer POE~\cite{li2021learning} & 0.651 & 0.816 & 0.642 & 0.604 & 0.703 & 0.565\\
            Transformer CPL hard~\cite{Wang_Jiang_Yin_Cheng_Ge_Gu_2023}& 0.669 & 0.817 & 0.662 & 0.668 & 0.759 & 0.633\\
            Transformer CPL soft~\cite{Wang_Jiang_Yin_Cheng_Ge_Gu_2023} & 0.641 & 0.793 & 0.616 & 0.652 & 0.726 & 0.612 \\
    
            \rowcolor{gray!15}
            Ours & \bf{0.690} & \bf{0.826} & \bf{0.674} & \bf{0.683} & \bf{0.774} & \bf{0.646}\\ \hline
        \end{tabular}
        }
        \vspace{-2mm}
        \label{tab:ordinalcomparison}
\end{table*}

\section{Experiments}
\noindent
{\bf Dataset.}
We used two datasets: a publicly available labeled ulcerative colitis dataset ({\bf LIMUC}~\cite{polat2022class}), and a private dataset collected from the {\it Kyoto Second Red Cross Hospital} ({\bf Private}). The LIMUC dataset includes 11,276 images collected from 564 patients with image-level annotation. The number of images in the bag ranges from 1 to 105, with an average size of 20 images per bag.
To simulate realistic scenarios, we assigned the label of the most severe part of the image within each bag to the bag itself.
The private dataset consists of 10,265 images collected from 388 patients with patient-level severity labels. The number of images in the bag ranges from 1 to 86, with various bags, and the average bag size is 26. 
Note that the image-level annotations were additionally made for evaluation. Only patient-level labels are available, and no image-level labels exist when applying our method in different domains or similar applications.



\noindent
{\bf Implementation Details.} 
We implemented our method by using PyTorch \cite{paszke2019pytorch}.
For the instance feature extractor $f$, we used ResNet18~\cite{he2016deep} pre-trained on the ImageNet dataset~\cite{deng2009imagenet}. The network was trained using the Adam optimizer \cite{adam}, with a learning rate of $3e-6$, $\mathrm{epoch}=1500$, a mini-batch size$=32$, and early stopping$=100$ patients. 
Additionally, to address the class imbalance in the data, oversampling based on the number of patient-level severity labels was performed during training.
Each method is evaluated with 5-fold cross-validation.

\noindent
{\bf Evaluation Metrics.}
We evaluated the performance of the proposed method based on three metrics: classification accuracy ({\bf Accuracy}), {\bf Macro-f1} for a robust metric against imbalance, and quadratic weighted kappa ({\bf Kappa})~\cite{polat2022class}, which assesses agreement between predicted and actual categorical labels, considering both the degree of error and ordinal relationships.

\subsection{Comparisons} 
\label{sec:comparison}


\begin{table*}[t]
    \def\@captype{table}
      \makeatother
        \centering
        \caption{{\bf Effectiveness of selective aggregated transformer on LIMUC and Private dataset.} 
         Best performances are bold.
         ``K-rank'' is K-rank algorithm.
        ``SAT'' is our selective aggregated transformer. 
        Transformer K-rank is a method that removes the selective aggregation module from our proposed method, which is different from K-rank MIL.}
        \vspace{-2mm}
        \scalebox{0.97}{
     \begin{tabular}{c|| cc ||ccc ||ccc} \hline
     \multirow{2}{*}{Method} & \multirow{2}{*}{K-rank} & \multirow{2}{*}{SAT} &    \multicolumn{3}{c||}{LIMUC} & \multicolumn{3}{c}{Private} \\
        & &&   Accuracy & Kappa & Macro-f1 & Accuracy & Kappa & Macro-f1\\ \hline \hline
        Transformer  &&& 0.665 & 
0.813 & 0.641 & 0.634 & 0.730 & 0.608\\
        Transformer K-rank &$\checkmark$&&0.640 & 0.807 & 0.620 & 0.670 & 0.762 & 0.630\\
        \rowcolor{gray!15}
        Ours &$\checkmark$&$\checkmark$ & \bf{0.690} & \bf{0.826} & \bf{0.674} & \bf{0.683} & \bf{0.774} & \bf{0.646}\\
        \hline        
        \end{tabular}
        }
        \label{tab:ablation1}
        \vspace{0mm}
\end{table*}

\noindent
{\bf Comparison with MIL Methods:}
We compared our method with ten MIL methods, including state-of-the-art methods.
1) to 4) ``Output+mean, max''~\cite{wang2018revisiting} and ``Feature+mean, max''~\cite{Ilse2020DeepMI} aggregate instance scores or features in a bag by using pooling operations.
5), 6) ``Feature+Attention''~\cite{ilse2018attention} and ``Transformer'' 
aggregate instance features through weighted summation using attention mechanisms.
7) ``DSMIL''~\cite{li2021dual}, 8) ``AdditiveTransMIL''~\cite{javed2022additive} and 9) ``IBMIL''~\cite{lin2023interventional} are the state-of-the-art MIL methods, which were designed for whole slide images (WSI) classification but the main idea can be applied for general MIL problems.
10) ``K-rank MIL''~\cite{schwab2022automatic} is a state-of-the-art MIL method for severity estimation of UC at the patient level, simply combining the K-rank algorithm~\cite{niu2016ordinal} in ordinal classification with the Output+max~\cite{wang2018revisiting} method in MIL.

Table \ref{tab:milcomparison} shows the comparison with the MIL methods.
Traditional output aggregation produced worse performances due to insufficient feature representation. Feature aggregation with simple pooling (mean and max) improved the performances.
``Feature+Attention'' and ``Transformer'' achieved superior performance than them by weighting instances with higher severity using attention mechanisms or self-attention. However, it is difficult for them to distinguish between neighboring severity classes.
In addition, our method outperformed the state-of-the-art MIL methods "DSMIL," "AdditiveTransMIL," and "IBMIL" because these methods are not designed for max severity estimation in ordinal MIL. Consequently, their aggregated features can be confusing at adjacent severity levels.

``K-rank MIL'', one of the most related works, achieved better Kappa and Macro-f1 than these state-of-the-art methods because this method is designed for ordinal classification MIL. However, this method simply introduced K-rank label estimation to MIL, and the aggregation is performed on output scores for each K-rank label estimation. Therefore, it is worse than a transformer-based method.
Our method achieved the best performance on all the metrics.
A method that introduces the K-rank algorithm to the transformer will be discussed in the ablation study.


\begin{figure}
      \centering
        \includegraphics[width=0.95\linewidth]{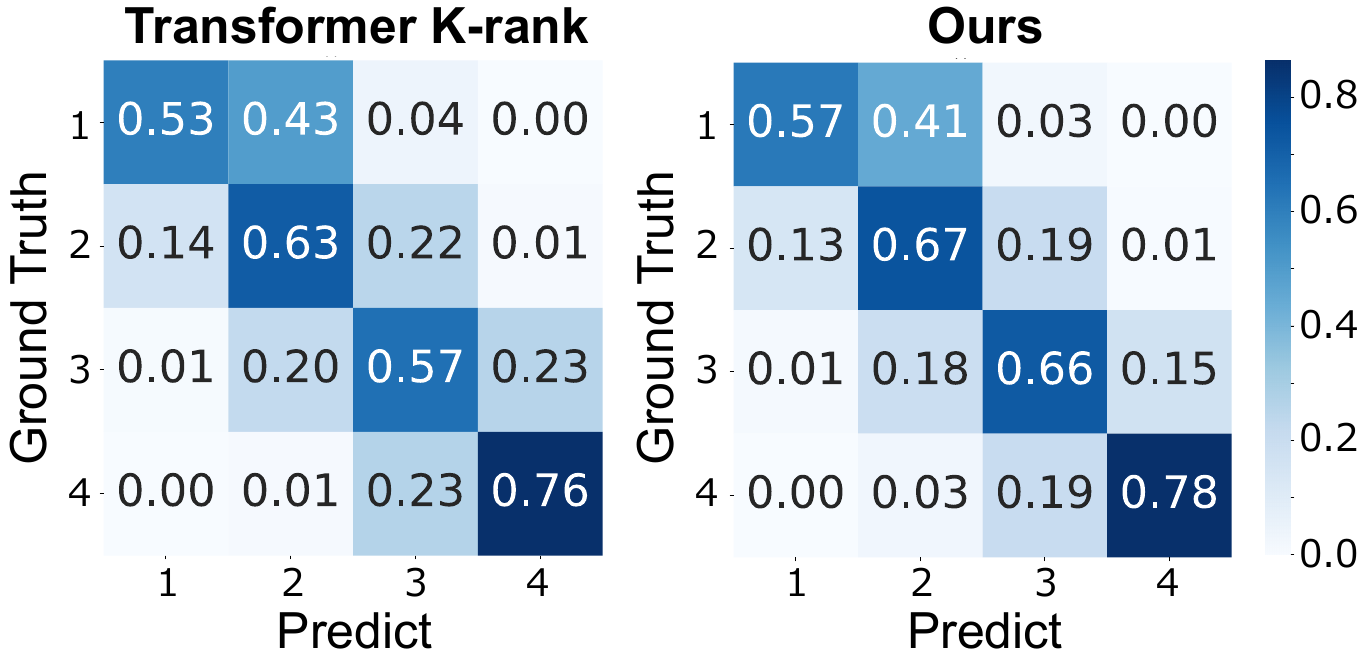}
        \vspace{-2mm}
        \caption{{\bf Confusion matrix} of Transformer K-rank (w/o selective aggregator token) and Ours using LIMUC. 
        }
        \label{fig:confusion}
        \vspace{-5mm}
\end{figure}

\noindent
{\bf Comparison with Ordinal Classification Methods:}
Since there are few methods that address ordinal classification in MIL, for comparison, we introduced five ordinal classification methods into the ``Transformer'', which is the second-best among MIL methods, following ours.
1) ``Transformer regression'', which outputs regression.
2) ``Transformer soft label''~\cite{diaz2019soft}, which utilizes soft labels that reflect ordinality.
3) ``Transformer POE''~\cite{li2021learning}, which obtains ordinal distribution by modeling data uncertainty.
4) ``Transformer CPL''~\cite{Wang_Jiang_Yin_Cheng_Ge_Gu_2023}, which is a state-of-the-art method for ordinal classification, enforces ordinal distribution on the feature space through Constrained Proxies Learning (CPL), which has two options (soft) and (hard). 

Table \ref{tab:ordinalcomparison} shows the comparison with ordinal classification methods.
These methods account for ordinal relationships of severity classes: greater loss is assigned for errors where the predicted class is farther from the true class compared to errors where the predicted class is closer, thereby considering the order. However, the performance improvement was limited. This is because their feature aggregation for max severity estimation in ordinal MIL has the same disadvantage as the other MIL methods discussed above.
Our method outperformed these MIL methods that are customized for ordinal classification.

\begin{figure}
      \centering
        \includegraphics[width=0.9\linewidth]{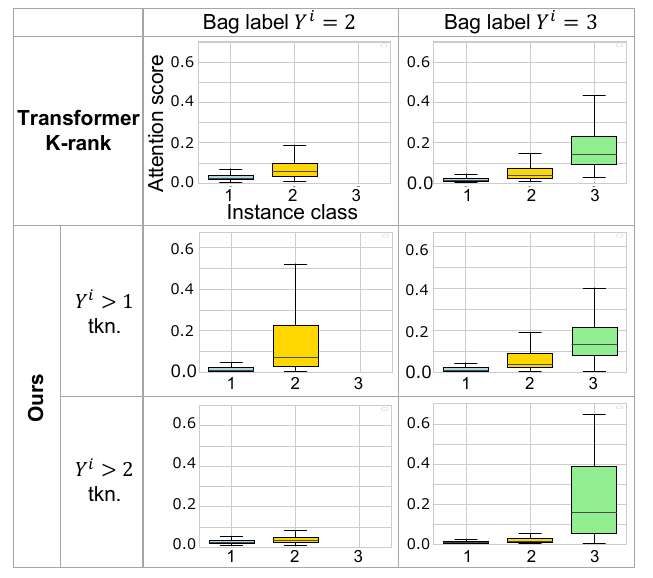}
        \vspace{-3mm}
        \caption{{\bf Distribution of attention weights} for each instance within the bag for class 2 and class 3 estimated by Ours and Transformer K-rank using LIMUC. The horizontal axis indicates the instance classes, and the vertical axis indicates the attention score. Note that our method produces multiple bag-level features for each bag using multiple tokens, while Transformer K-rank produces a single feature.
        }
        \vspace{-3mm}
        \label{fig:boxplot}
\end{figure}


\subsection{Effectiveness of Selective Aggregated Transformer} 
We conducted an ablation study to investigate the effectiveness of aggregation,  focusing on the classification of adjacent classes using the Selective Aggregated Transformer.
We compared our method with Transformer-based MIL (``Transformer'') and ``Transformer K-rank'', which is a method that combines the K-rank algorithm~\cite{cao2020rank,niu2016ordinal} and the Transformer method, removing the selective aggregation module from our proposed method.

Table~\ref{tab:ablation1} shows the performance of each comparison.
Introducing ordinal classification to the Transformer did not improve performance. This is because the feature aggregator, by a single token, does not have the ability to distinguish K-rank multi-labels.  
Our selective aggregated transformer improved the performance, indicating the proposed method's effectiveness.

Figure~\ref{fig:confusion} shows the confusion matrix for ``Transformer K-rank'' and our method.
This result confirms that the values of all the diagonal elements have improved and that the misclassifications between adjacent classes have decreased.
This result indicates that the proposed method, by introducing selective aggregator tokens, is improving the classification performance of adjacent classes.

\begin{figure*}[t]
      \centering
        \includegraphics[width=1.0\linewidth]{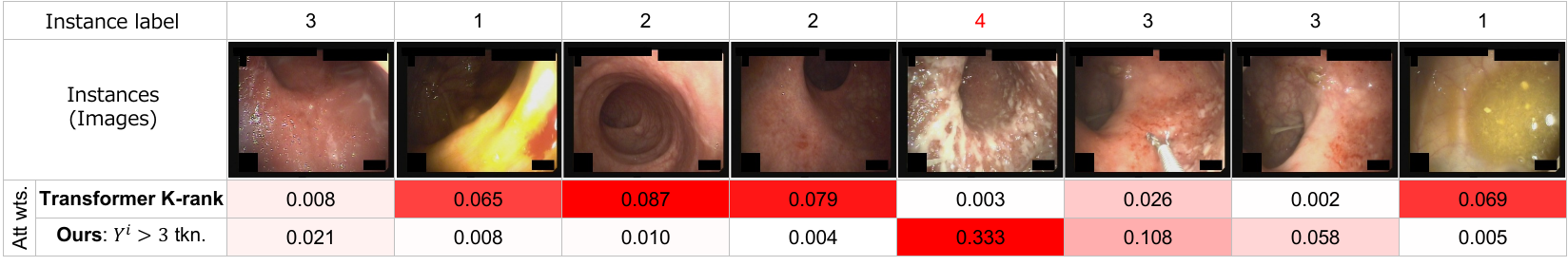}
        \vspace{-6mm}
        \caption{{\bf Example of images and their attention scores in LIMUC}, which are estimated by the single token of Transformer K-rank and the $Y^i > 3$ token of the proposed selective aggregated Transformer.
        The color indicates that higher attention is represented by a stronger red hue.
        }
        \vspace{-4mm}
        \label{fig:qualitative}
\end{figure*}

\begin{figure}
      \centering
        \includegraphics[width=1.0\linewidth]{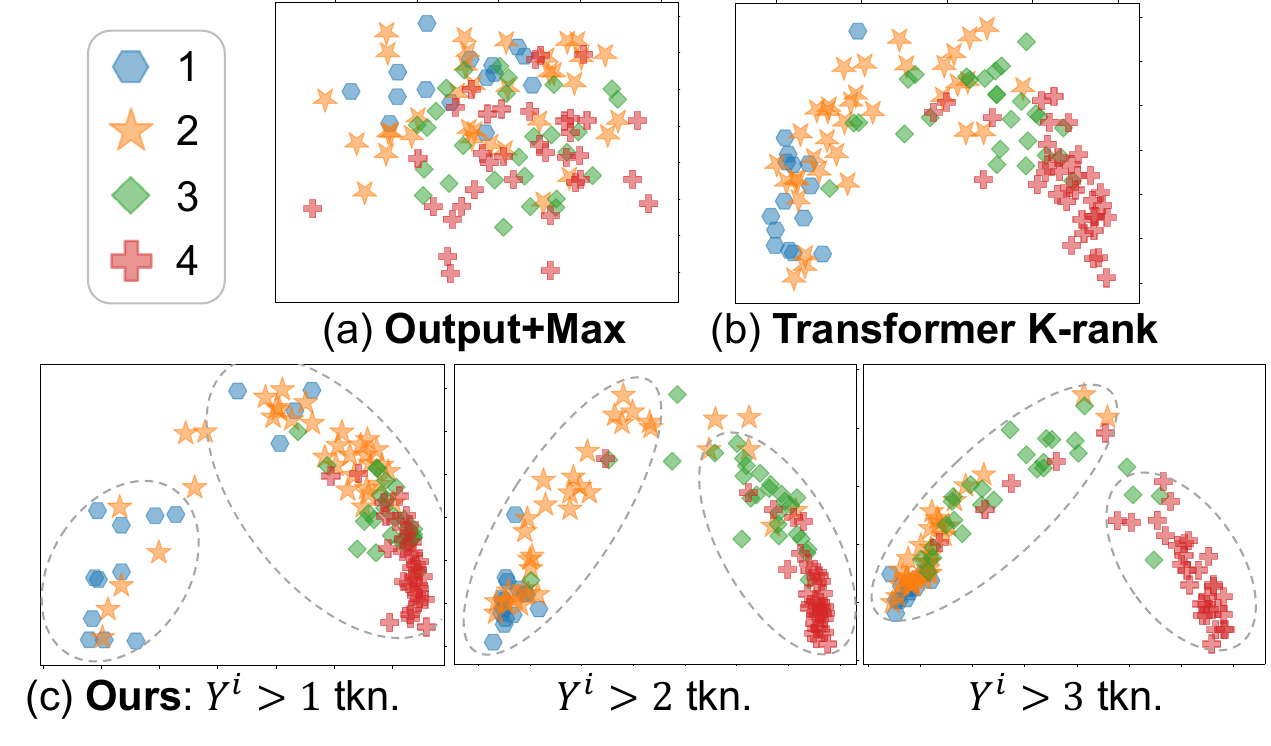}
        \vspace{-5mm}
        \caption{{\bf Visualization of features spaces using LIMUC.} 
        Our method produces $K-1$ features for each bag by multiple tokens, while others only produce a single one. Each token successfully produces discriminative bag-level features between adjacent classes. 
        }
        \vspace{-4mm}
        \label{fig:feat_vis}
\end{figure}

To demonstrate the effectiveness of our selective aggregation module for attention estimation, we visualize the distributions of attention scores estimated by ours and ``Transformer K-rank''.
Figure~\ref{fig:boxplot} illustrates the attention score distributions for each instance within the bag for classes 2 and 3, where instances with the top 50\% weights that influence estimation are visualized. ``Transformer K-rank'', the distributions of attention scores estimated by the single token are visualized. In the proposed method, the distributions of attention scores estimated by the $\{Y^i > 1\}$ (indicating $\bm{t}_1$) and $\{Y^i > 2\}$ (indicating $\bm{t}_2$) tokens for classifying class 2 and class 3 are visualized. 

The result shows that the ``Transformer K-rank'' estimates slightly high attention to class 3; however, attention to instances with class 2 is also a bit higher than 1. In particular, the distributions of instances with severity 1 and 2 in bag 2 overlap. Since this method only uses the attention obtained by a single token for all classes of bags, appropriate attention assignments should be performed for all types of bags. However, it makes it difficult to distinguish all classes.

In contrast, our selective tokens can aggregate discriminative features between adjacent classes.
The role of token $Y^i > 2$ is to distinguish between instances with severity 3 and lower severities. Consequently, it is expected to give high attention only to instances with severity 3 in bags with severity 3, while giving low attention to other instances in both bags. The results show that, as expected, the tokens $Y^i > 1$ and $Y^i > 2$ successfully distinguish between adjacent classes, respectively.

Figure~\ref{fig:qualitative} shows an example of images and their attention scores estimated by the single token of ``Transformer K-rank'' and the $Y^i > 3$ token of the proposed selective aggregated transformer.
``Transformer K-rank'' incorrectly estimates high attention scores for instances whose labels are less than 4. Additionally, it estimates similar attention scores for instances across multiple classes. This leads to the production of un-discriminative bag-level features.
In contrast, the proposed method estimates high attention for the most severe instances and relatively low attention for instances from other classes.

To demonstrate the ability of each selective aggregator token to obtain discriminative features between adjacent classes, we visualize the bag-level feature distributions obtained by three methods, as shown in Figure~\ref{fig:feat_vis}: (a) ``Output+Max'', (b) ``Transformer  K-rank'', and (c) features of each selective aggregator token ${Y^i > k},(k=1,2,3)$ in our proposed method. Here, ``Output+max'' does not aggregate instance features, so it visualizes the features of the selected instance using the max operation.
``Ouput+Max'' and ``Transformer K-rank'' only have a single feature space. In contrast, our method has $K-1$ feature spaces extracted by each selective aggregator token.

Although ``Output+Max'' is the most straightforward approach for our task, it failed to separate the features selected by the max operation between different classes. This result shows the poor ability of ``Output+Max'' to learn representation.
``Transformer K-rank'' improved the bag-level feature representation. However, the distributions of neighbor classes partially overlap. It shows the limitation of aggregation by a single token.
In contrast, our method successfully separates the distributions of adjacent classes ($Y^i>k$ or below) by the three selective aggregator tokens $\bm{t}_1, \bm{t}_2, \bm{t}_3$ ($Y^i>1, Y^i>2, Y^i>3$). 
As expected, the $k$-th token well separates distributions between classes $k-1$ and $k$.


\subsection{Comparison with an Image-level and Our Patient-level Method.} 
We also compared our method with two simple supervised methods that use image-level labels in a patient-level (bag-level) classification task. ``CL'' is a standard classification method that uses cross-entropy loss. ``K-rank''~\cite{cao2020rank,niu2016ordinal} is an ordinal classification method for classifying individual images. 
In training, we train instance-level classifiers using supervised labels, where we used the same network architecture for the image feature extractor for both methods.
During inference in both methods, the severities of individual images are estimated, and the patient-level label is determined by taking the maximum severity.
In contrast, our method trains a patient-level classifier directly in the MIL setting (only using patient labels).
Note that the image-level labels have been added for evaluation. In the real world, patient-level labels are only available in clinical.

\begin{table}[t]
    \def\@captype{table}
      \makeatother
        \centering
        \caption{{\bf Comparison with an image-level and our patient-level method in terms of Kappa.}
        This result demonstrates that our method can effectively train the network for patient-level diagnosis by only using existing clinical records, without requiring additional costly annotations.
        }    
        \vspace{-2mm}
        \scalebox{0.97}{
     \begin{tabular}{c|| c || c|c} \hline
     Method & Image label  & LIMUC & Private\\
        \hline \hline
        CL &$\checkmark$&0.782 & 0.755\\
        K-rank &$\checkmark$&0.818 & 0.760\\
        \rowcolor{gray!15}
        Ours &&\bf{0.826} & \bf{0.774} \\
        \hline  
        \end{tabular}
        }
        \label{tab:image_level}
\end{table}

Table~\ref{tab:image_level} shows the bag-level performance metric Kappa of these three methods.
Although our method only uses patient-level labels, it surprisingly outperformed the supervised methods that require image-level labels.
The supervised learning with output max did not work well because a misclassification for a maximum score directly affects the bag-level classification. 
This result demonstrates that our method can effectively train the network for patient-level diagnosis by only using existing information in clinical records without additional annotations.

\section{Conclusion}
We propose a method for patient-level severity estimation by transformer with selective aggregator tokens that can effectively use pre-recorded diagnosis reports without additional annotation. 
This task is one of the ordinal multiple-instance learning frameworks. 
We introduced ordinal classification techniques to the multiple-instance learning method. 
In addition, we extract multiple bag-level features for each classification boundary by transformer with multiple selective aggregator tokens.
Our selective aggregator tokens allow instances to be selected and aggregated the selected instances for each classification boundary.
The experimental results have proven the superior performance of the proposed method. 
Furthermore, we demonstrate the efficiency of our method in real clinical settings.


\noindent
{\bf Acknowledgements}: This work was supported by  SIP-JPJ012425, JSPS KAKENHI Grant JP23K18509, JSPS KAKENHI, JP24KJ2205, and JST ACT-X, Grant Number JPMJAX21AK. \\

\vspace{-8mm}
{\small
\bibliographystyle{ieee_fullname}
\bibliography{egbib}

\begin{thebibliography}{10}\itemsep=-1pt

\bibitem{bi2021local}
Qi Bi, Shuang Yu, Wei Ji, Cheng Bian, Lijun Gong, Hanruo Liu, Kai Ma, and
  Yefeng Zheng.
\newblock Local-global dual perception based deep multiple instance learning
  for retinal disease classification.
\newblock In {\em Medical Image Computing and Computer Assisted
  Intervention--MICCAI 2021: 24th International Conference, Strasbourg, France,
  September 27--October 1, 2021, Proceedings, Part VIII 24}, pages 55--64.
  Springer, 2021.

\bibitem{cao2020rank}
Wenzhi Cao, Vahid Mirjalili, and Sebastian Raschka.
\newblock Rank consistent ordinal regression for neural networks with
  application to age estimation.
\newblock {\em Pattern Recognition Letters}, 140:325--331, 2020.

\bibitem{deng2009imagenet}
Jia Deng, Wei Dong, Richard Socher, Li-Jia Li, Kai Li, and Li Fei-Fei.
\newblock Imagenet: A large-scale hierarchical image database.
\newblock In {\em 2009 IEEE conference on computer vision and pattern
  recognition}, pages 248--255. Ieee, 2009.

\bibitem{diaz2019soft}
Raul Diaz and Amit Marathe.
\newblock Soft labels for ordinal regression.
\newblock In {\em the IEEE/CVF conference on computer vision and pattern
  recognition}, pages 4738--4747, 2019.

\bibitem{guo2013joint}
Guodong Guo and Guowang Mu.
\newblock Joint estimation of age, gender and ethnicity: Cca vs. pls.
\newblock In {\em 2013 10th IEEE international conference and workshops on
  automatic face and gesture recognition (FG)}, pages 1--6. IEEE, 2013.

\bibitem{he2016deep}
Kaiming He, Xiangyu Zhang, Shaoqing Ren, and Jian Sun.
\newblock Deep residual learning for image recognition.
\newblock In {\em the IEEE conference on computer vision and pattern
  recognition}, pages 770--778, 2016.

\bibitem{ilse2018attention}
Maximilian Ilse, Jakub Tomczak, and Max Welling.
\newblock Attention-based deep multiple instance learning.
\newblock In {\em International conference on machine learning}, pages
  2127--2136. PMLR, 2018.

\bibitem{Ilse2020DeepMI}
Maximilian Ilse, Jakub~M. Tomczak, and Max Welling.
\newblock Deep multiple instance learning for digital histopathology.
\newblock In {\em Handbook of Medical Image Computing and Computer Assisted
  Intervention}, pages 521--546. Academic Press, 2020.

\bibitem{javed2022additive}
Syed~Ashar Javed, Dinkar Juyal, Harshith Padigela, Amaro Taylor-Weiner, Limin
  Yu, and Aaditya Prakash.
\newblock Additive mil: intrinsically interpretable multiple instance learning
  for pathology.
\newblock {\em Advances in Neural Information Processing Systems},
  35:20689--20702, 2022.

\bibitem{kadota2022automatic}
Takeaki Kadota, Kentaro Abe, Ryoma Bise, Takuji Kawamura, Naokuni Sakiyama,
  Kiyohito Tanaka, and Seiichi Uchida.
\newblock Automatic estimation of ulcerative colitis severity by learning to
  rank with calibration.
\newblock {\em IEEE Access}, 10:25688--25695, 2022.

\bibitem{kadota2022deep}
Takeaki Kadota, Hideaki Hayashi, Ryoma Bise, Kiyohito Tanaka, and Seiichi
  Uchida.
\newblock Deep bayesian active-learning-to-rank for endoscopic image data.
\newblock In {\em Annual Conference on Medical Image Understanding and
  Analysis}, pages 609--622. Springer, 2022.

\bibitem{li2021dual}
Bin Li, Yin Li, and Kevin~W Eliceiri.
\newblock Dual-stream multiple instance learning network for whole slide image
  classification with self-supervised contrastive learning.
\newblock In {\em the IEEE/CVF conference on computer vision and pattern
  recognition}, pages 14318--14328, 2021.

\bibitem{li2021learning}
Wanhua Li, Xiaoke Huang, Jiwen Lu, Jianjiang Feng, and Jie Zhou.
\newblock Learning probabilistic ordinal embeddings for uncertainty-aware
  regression.
\newblock In {\em the IEEE/CVF conference on computer vision and pattern
  recognition}, pages 13896--13905, 2021.

\bibitem{lin2023interventional}
Tiancheng Lin, Zhimiao Yu, Hongyu Hu, Yi Xu, and Chang-Wen Chen.
\newblock Interventional bag multi-instance learning on whole-slide
  pathological images.
\newblock In {\em the IEEE/CVF Conference on Computer Vision and Pattern
  Recognition}, pages 19830--19839, 2023.

\bibitem{niu2016ordinal}
Zhenxing Niu, Mo Zhou, Le Wang, Xinbo Gao, and Gang Hua.
\newblock Ordinal regression with multiple output cnn for age estimation.
\newblock In {\em the IEEE conference on computer vision and pattern
  recognition}, pages 4920--4928, 2016.

\bibitem{adam}
Kingma~Diederik P and Ba Jimmy.
\newblock Adam: A method for stochastic optimization.
\newblock In {\em arXiv}, 2014.

\bibitem{paszke2019pytorch}
Adam Paszke, Sam Gross, Francisco Massa, Adam Lerer, James Bradbury, Gregory
  Chanan, Trevor Killeen, Zeming Lin, Natalia Gimelshein, Luca Antiga, et~al.
\newblock Pytorch: An imperative style, high-performance deep learning library.
\newblock {\em Advances in neural information processing systems}, 32, 2019.

\bibitem{Pinheiro2015}
P.~O. Pinheiro and R. Collobert.
\newblock From image-level to pixel-level labeling with convolutional networks.
\newblock In {\em Computer Vision and Pattern Recognition}, pages 1713--1721,
  2015.

\bibitem{polat2022class}
Gorkem Polat, Ilkay Ergenc, Haluk~Tarik Kani, Yesim~Ozen Alahdab, Ozlen Atug,
  and Alptekin Temizel.
\newblock Class distance weighted cross-entropy loss for ulcerative colitis
  severity estimation.
\newblock In {\em Annual Conference on Medical Image Understanding and
  Analysis}, pages 157--171. Springer, 2022.

\bibitem{ramon2000multi}
Jan Ramon and Luc De~Raedt.
\newblock Multi instance neural networks.
\newblock In {\em International Conference on Machine Learning workshop}, pages
  53--60, 2000.

\bibitem{schwab2022automatic}
Evan Schwab, Gabriela~Oana Cula, Kristopher Standish, Stephen~SF Yip,
  Aleksandar Stojmirovic, Louis Ghanem, and Christel Chehoud.
\newblock Automatic estimation of ulcerative colitis severity from endoscopy
  videos using ordinal multi-instance learning.
\newblock {\em Computer Methods in Biomechanics and Biomedical Engineering:
  Imaging \& Visualization}, 10(4):425--433, 2022.

\bibitem{shao2021transmil}
Zhuchen Shao, Hao Bian, Yang Chen, Yifeng Wang, Jian Zhang, Xiangyang Ji,
  et~al.
\newblock Transmil: Transformer based correlated multiple instance learning for
  whole slide image classification.
\newblock {\em Advances in neural information processing systems},
  34:2136--2147, 2021.

\bibitem{stidham2019performance}
Ryan~W Stidham, Wenshuo Liu, Shrinivas Bishu, Michael~D Rice, Peter~DR Higgins,
  Ji Zhu, Brahmajee~K Nallamothu, and Akbar~K Waljee.
\newblock Performance of a deep learning model vs human reviewers in grading
  endoscopic disease severity of patients with ulcerative colitis.
\newblock {\em JAMA network open}, 2(5):e193963--e193963, 2019.

\bibitem{takenaka2020development}
Kento Takenaka, Kazuo Ohtsuka, Toshimitsu Fujii, Mariko Negi, Kohei Suzuki,
  Hiromichi Shimizu, Shiori Oshima, Shintaro Akiyama, Maiko Motobayashi,
  Masakazu Nagahori, et~al.
\newblock Development and validation of a deep neural network for accurate
  evaluation of endoscopic images from patients with ulcerative colitis.
\newblock {\em Gastroenterology}, 158(8):2150--2157, 2020.

\bibitem{isbi_takezaki}
Shumpei Takezaki, Kiyohito Tanaka, Seiichi Uchida, and Takeaki Kadota.
\newblock {Disease Severity Regression with Continuous Data Augmentation}.
\newblock In {\em International Symposium on Biomedical Imaging}, pages 1--5,
  2023.

\bibitem{Wang_Jiang_Yin_Cheng_Ge_Gu_2023}
Cong Wang, Zhiwei Jiang, Yafeng Yin, Zifeng Cheng, Shiping Ge, and Qing Gu.
\newblock Controlling class layout for deep ordinal classification via
  constrained proxies learning.
\newblock {\em Proceedings of the Association for the Advancement of Artificial
  Intelligence Conference on Artificial Intelligence}, 37(2):2483--2491, Jun.
  2023.

\bibitem{wang2018revisiting}
Xinggang Wang, Yongluan Yan, Peng Tang, Xiang Bai, and Wenyu Liu.
\newblock Revisiting multiple instance neural networks.
\newblock {\em Pattern Recognition}, 74:15--24, 2018.

\bibitem{yu2021mil}
Shuang Yu, Kai Ma, Qi Bi, Cheng Bian, Munan Ning, Nanjun He, Yuexiang Li,
  Hanruo Liu, and Yefeng Zheng.
\newblock Mil-vt: Multiple instance learning enhanced vision transformer for
  fundus image classification.
\newblock In {\em Medical Image Computing and Computer Assisted
  Intervention--MICCAI 2021: 24th International Conference, Strasbourg, France,
  September 27--October 1, 2021, Proceedings, Part VIII 24}, pages 45--54.
  Springer, 2021.

\end{thebibliography}
}

\end{document}